\documentclass[10pt,journal,compsoc]{IEEEtran}
\usepackage{graphicx}
\usepackage{amsmath}
\usepackage{amssymb}
\usepackage{booktabs}
\usepackage{times}
\usepackage{epsfig}
\usepackage{multirow}
\usepackage{mathtools}
\usepackage{xcolor}
\usepackage{flexisym}
\usepackage{adjustbox}
\usepackage{diagbox}
\usepackage{makecell}
\usepackage{hyperref}

\def\etal{\emph{et al.}}
\def\eg{\emph{e.g.}}
\def\ie{\emph{i.e.}}

\def\wrt{\emph{w.r.t.}}

\definecolor{cfgreen}{rgb}{0.0, 0.42, 0.24}

\DeclarePairedDelimiter{\norm}{\lVert}{\rVert}

\ifCLASSOPTIONcompsoc
  \usepackage[nocompress]{cite}
\else
  \usepackage{cite}
\fi
\ifCLASSINFOpdf
\else
\fi
\begin{document}
%
\title{Generating Multiple 4D Expression Transitions by Learning Face Landmark Trajectories}
%
%
%
%



\author{Naima~Otberdout*,
        Claudio~Ferrari*,
        Mohamed~Daoudi,
        Stefano~Berretti,
        Alberto~Del~Bimbo,

\IEEEcompsocitemizethanks{\IEEEcompsocthanksitem * Equal contributions.\\ \IEEEcompsocthanksitem N. Otberdout is with Ai movement - University Mohammed VI Polytechnic, Rabat, Morocco, E-mail: naima.otberdout@um6p.ma\\
\IEEEcompsocthanksitem C. Ferrari is with the Department of Architecture and Engineering University of Parma, Italy, E-mail: claudio.ferrari2@unipr.it\\
\IEEEcompsocthanksitem
M. Daoudi is with  Univ. Lille, CNRS, Centrale Lille, Institut Mines-Télécom, UMR 9189 CRIStAL, F-59000 Lille, France and IMT Nord Europe, Institut Mines-Télécom, Univ. Lille, Centre for Digital Systems, F-59000 Lille, France, E-mail: mohamed.daoudi@imt-nord-europe.fr\\
\IEEEcompsocthanksitem S. Berretti and A. Del Bimbo are with Media Integration and Communication Center (MICC), Univ. of Florence, Italy. E-mail: \{stefano.berretti, alberto.delbimbo\}@unifi.it}

\thanks{Manuscript received July 14, 2022; revised ????}}

%
%

\markboth{Journal of \LaTeX\ Class Files,~Vol.~14, No.~8, July~2022}%
{Otberdout \MakeLowercase{\textit{et al.}}: Generating Multiple 4D Expression Transitions
by Learning Face Landmark Trajectories}
%



\IEEEtitleabstractindextext{%
\begin{abstract}
In this work, we address the problem of 4D facial expressions generation. This is usually addressed by animating a neutral 3D face to reach an expression peak, and then get back to the neutral state. In the real world though, people show more complex expressions, and switch from one expression to another. We thus propose a new model that generates transitions between different expressions, and synthesizes long and composed 4D expressions. This involves three sub-problems: \textit{(i)} modeling the temporal dynamics of expressions, \textit{(ii)} learning transitions between them, and \textit{(iii)} deforming a generic mesh. We propose to encode the temporal evolution of expressions using the motion of a set of 3D landmarks, that we learn to generate by training a manifold-valued GAN (Motion3DGAN). To allow the generation of composed expressions, this model accepts two labels encoding the starting and the ending expressions. The final sequence of meshes is generated by a Sparse2Dense mesh Decoder (S2D-Dec) that maps the landmark displacements to a dense, per-vertex displacement of a known mesh topology. By explicitly working with motion trajectories, the model is totally independent from the identity. Extensive experiments on five public datasets show that our proposed approach brings significant improvements with respect to previous solutions, while retaining good generalization to unseen data. 
\end{abstract}

}

\maketitle

\IEEEdisplaynontitleabstractindextext

%
\IEEEpeerreviewmaketitle

\IEEEraisesectionheading{\section{Introduction}\label{sec:intro}}

%
%
%
%
\IEEEPARstart{G}{enerating} dynamic 3D (4D) face models is the task of synthesizing realistic 3D face instances that dynamically evolve across time with varying expressions or speech-related movements, while keeping the same identity.
This can be useful in a wide range of graphics applications, spanning from 3D face modeling to augmented and virtual reality for animated films and computer games. While recent advances in generative neural networks have made possible the development of effective solutions that operate on 2D images~\cite{Fan-AAAI:2019, OtberdoutPAMI2020}, the literature on the problem of generating facial animation in 3D is still quite limited, with few examples available~\cite{PotamiasECCV2020, Otberdout_2022_CVPR} 
\begin{figure}[!t]
\centering 
\includegraphics[width=\linewidth]{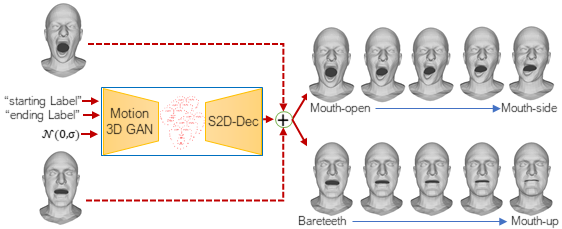}
\caption{\textbf{3D dynamic facial expression generation}: A GAN generates the motion of 3D landmarks from a pair of expression labels, \ie, one starting and one ending labels and noise; A decoder expands the animation from the landmarks to a dense mesh, while keeping the identity of a neutral 3D face.}
\label{fig:overview}
\end{figure}

Performing faithful and accurate 3D facial animations requires addressing some major challenges, in terms both of 3D face modeling, and temporal dynamics. Related to the former, as we wish to animate a 3D face of an individual, its identity should be maintained across time. Also, the applied dynamic deformation should be controllable, corresponding to a specific expression/motion, and should be applicable to any 3D face. Incidentally, these are major challenges in 3D face modeling, which require disentangling structural face elements related to the identity, \eg, nose or jaw shape, from deformations related to the movable face parts, \eg, mouth opening/closing. Modeling the temporal dynamics of expressions, instead, gives rise to other challenges. First, expressions are very personal, and different individuals might perform the same expressions differently; for example, it is difficult to find two people laughing in the same exact way. So, for the sake of realism, one should be able to generate diverse motions even for the same expression. 
Furthermore, dynamic expressions have been always standardized in a way that they are supposed to start from a neutral configuration, reach (onset) a peak of the expression (apex), and then get back (offset) to neutral~\cite{zhang2013high, cosker2011facs,COMA:ECCV18}. However, this is not the case in the real world, where people might switch from one expression to another dynamically, so that a sequence could start from or end to a non-neutral configuration. This implies considering the sequence as a transition between two expressions.

Some previous works tackled the problem of neutral-to-apex generation by capturing the facial expression of a subject frame-by-frame and transferring it to a target model~\cite{cao-tog:2014}. However, in this case, the temporal evolution is neglected, so the problem reduces to transferring a tracked expression to a neutral 3D face. Some other works animated a 3D face mesh given an arbitrary speech signal and a static 3D face mesh as input~\cite{Cudeiro-cvpr:2019, Karras-tog:2017}, some also considering additional emotional labels~\cite{ji2022eamm, zhao2022emotionally, eskimez2021speech}. Also in this case though, the temporal evolution is guided by an external input, similar to a tracked expression. 
Differently, here we are interested in animating a face just starting from a generic 3D face and a pair of expression labels, \ie, a starting and an ending label.

In our solution, which is illustrated in Figure~\ref{fig:overview}, the temporal evolution and the mesh deformation are decoupled and modeled separately in two network architectures. A manifold-valued GAN (\textit{Motion3DGAN}) accounts for the expression dynamics by generating a temporally consistent motion of 3D landmarks corresponding to a specified transition between two expressions from a noise signal. The landmarks motion is encoded using the \textit{Square Root Velocity Function} (SRVF), and compactly represented as a point on a hypersphere. The novel characteristic of this network is that, differently from all the previous literature, it can generate motions that do not necessarily start from a neutral configuration. In particular, the sequence to be modeled is specified by two labels, encoding the starting face configuration, and the ending one. In this way, we are able to generate \textit{(i)} arbitrarily long, and \textit{(ii)} composed motions of landmarks.
Then, a Sparse2Dense mesh Decoder (\textit{S2D-Dec}) generates a dense 3D face guided by the landmarks motion for each frame of the sequence.Ultimately, the two networks allow us to generate a dynamic sequence of 3D faces performing a dynamic transitions between expressions. To effectively disentangle identity and expression components, the landmarks motion is represented as a per-frame displacement (motion) from the starting configuration. Instead of directly generating a mesh, the S2D-Dec expands the landmarks displacement to a dense, per-vertex displacement, which is finally used to deform the neutral mesh. We thus train the decoder to learn how the displacement of a sparse set of points influences the displacement of the whole face surface. This has the advantage that structural face parts, \eg, nose or forehead, which are not influenced by facial expressions are ignored, helping in maintaining the identity traits stable. Furthermore, the network can focus on learning expressions at a fine-grained level of detail, and generalize to unseen identities.

In summary, the main contributions of our work are: 
\emph{(i)} we propose an original method to generate dynamic sequences of 3D expressive scans given a 3D face mesh and a pair of expression labels representing, respectively, the starting and ending expression of the sequence. This is the first solution that can generate smooth transitions in 3D between two generic expression labels, while works in the literature are constrained to the neutral-to-apex transition. Our approach can generate strong and diverse expression sequences, with high generalization ability to unseen identities and expressions. This has been obtained by adapting the GAN architecture proposed in~\cite{Otberdout_2022_CVPR} for accepting and learning from two labels. Doing so demanded for a dataset including transitions between expressions. Given that such dataset does not exist, we \emph{(ii)} defined a data augmentation strategy specific for 4D expression sequences, which is based on the SRVF encoding; 
finally, \emph{(iii)} we exploit the above characteristic of our model to generate concatenated sequences of expression transitions. An overall sequence can start from an expression $A$, change to an expression $B$, then to expressions $C$ and $D$. This is modeled as a combined generation from the pair $A-B$ to $B-C$ and $C-D$. We prove the expressions generated in the subsequent stages of the generation, \ie, $B-C$ and $C-D$, do not diverge though they are generated starting from the synthetic model generated at the end of the previous transition.

The rest of the paper is organized as follows: In Section~\ref{sect:related-work}, we summarize the works in~ the literature that are closer to our proposed solution; In Section~\ref{sect:proposed-method}, we introduce the proposed solution for generating the temporal dynamics of facial landmarks and to derive a dense mesh from them; A comprehensive experimental evaluation of our approach is presented in Section~\ref{sect:results}; Finally, conclusions and future work directions are given in Section~\ref{sect:conclusions}.

\section{Related Work}\label{sect:related-work}
Our work is related to methods for \emph{(a)} 3D face modeling, \emph{(b)} facial expression generation guided by landmarks, and \emph{(c)} dynamic generation of 3D faces, \ie, 4D face generation. Below, we summarize works in these three areas that are relevant for our proposal.

\subsection*{3D face modeling} 
The 3D Morphable face Model (3DMM) as originally proposed in~\cite{blanz1999morphable} is the most popular solution for modeling 3D faces. 
The original model and its variants~\cite{Booth-cvpr:2016, Paysan-avss:2009, brunton2014multilinear, neumann2013sparse, luthi2017gaussian, ferrari2021sparse, ferrari2015dictionary} capture face shape variations both for identity and expression based on linear formulations, thus incurring in limited modeling capabilities. For this reason, non-linear encoder-decoder architectures are attracting more and more attention. This comes at the cost of reformulating convolution and pooling/unpooling like operations on the irregular mesh support~\cite{Bronstein17, Litany-cvpr:2018, Verma-cvpr:2018}. For example, Ranjan~\etal~\cite{COMA:ECCV18} proposed an auto-encoder architecture that builds upon newly defined spectral convolution operators, and pooling operations to down-/up-sample the mesh. Bouritsas~\etal~\cite{Bouritsas:ICCV2019} improved upon the above by proposing a graph convolutional operator enforcing consistent local orderings on the vertices of the graph through the \textit{spiral operator}~\cite{lim2018_correspondence_learning}. Despite their impressive modeling precision, a recent work~\cite{ferrari2021sparse} showed that they heavily suffer from poor generalization to unseen identities. This limits their practical use in tasks such as face fitting or expression transfer. We finally mention that other approaches do exist to learn generative 3D face models, such as~\cite{Abrevaya-wacv:2018, Moschoglou-ijcv:2020}. However, instead of dealing with meshes they use alternative representations for 3D data, such as depth images or UV-maps.

To overcome the above limitation, we go beyond self-reconstruction and propose a mesh decoder that, differently from previous models, learns expression-specific mesh deformations from a sparse set of landmark displacements.

\subsection* {Facial expression generation guided by landmarks}
Recent advances in neural networks made facial landmark detection reliable and accurate both in 2D~\cite{Chen-NEURIPS:2019, Dong-tpami:2020, Wan-TNNLS:2021} and 3D~\cite{gilani2017deep, zhu2017face}. Landmarks and their motion are a viable way to account for facial deformations as they reduce the complexity of the visual data, and have been commonly used in several 3D face related tasks, \eg, reconstruction~\cite{ferrari2015dictionary, FLAME:SiggraphAsia2017} or reenactment~\cite{ferrari2018rendering, garrido2014automatic}. Despite some effort was put in developing landmark-free solutions for 3D face modeling~\cite{chang2018expnet, chang2017faceposenet, gecer2019ganfit}, some recent works investigated their use to model the dynamics of expressions. Wang~\etal~\cite{WangCVPR2018} proposed a framework that decouples facial expression dynamics, encoded into landmarks, and face appearance using a conditional recurrent network. Otberdout~\etal~\cite{OtberdoutPAMI2020} proposed an approach for generating videos of the six basic expressions given a neutral face image. The geometry is captured by modeling the motion of landmarks with a GAN that learns the distribution of expression dynamics. 

These methods demonstrated the potential of using landmarks to model the dynamics of expressions and generate 2D videos. In our work, we instead tackle the problem of modeling the dynamics in 3D, exploring the use of 3D landmarks motion to both model the temporal evolution of expressions and animate a 3D face.

\subsection*{4D face generation} 
While many researchers tackled the problem of 3D mesh deformation, the task of 3D facial motion synthesis is yet more challenging. A few studies addressed this issue by exploiting audio features~\cite{Karras-tog:2017, Zeng-ACMMM:2020}, speech signals~\cite{Cudeiro-cvpr:2019} or tracked facial expressions~\cite{cao-tog:2014} to generate facial motions. However, none of these explicitly models the temporal dynamics. 

The work in~\cite{PotamiasECCV2020} first addressed the problem of dynamic 3D expression generation. In that framework, the motion dynamics is modeled with a temporal encoder based on an LSTM, which produces a per-frame latent code starting from a per-frame expression label. The codes are then fed to a mesh decoder that, similarly to our approach, generates a per-vertex displacement that is summed to a neutral 3D face to obtain the expressive meshes. Despite the promising results reported in~\cite{PotamiasECCV2020}, we identified some limitations in this solution. First, the LSTM is deterministic, and for a given label the exact same displacements are generated. Our solution instead achieves diversity in the output sequences by generating from noise. Moreover, in~\cite{PotamiasECCV2020} the mesh decoder generates the displacements from the latent codes, making it dependent from the temporal encoder. In our solution, the motion dynamics and mesh displacement generation are decoupled, using landmarks to link the two modules. The S2D-Dec is thus independent from Motion3DGAN, and can be used to generate static meshes as well given an arbitrary set of 3D landmarks as input. This permits us to use the decoder for other tasks such as expression/speech transfer. Finally, as pointed out in~\cite{PotamiasECCV2020}, the model cannot perform extreme variations well. Using landmarks allowed us to define a novel reconstruction loss that weighs the error of each vertex with respect to its distance from the landmarks, encouraging accurate modeling of the movable parts. Thanks to this, we are capable of accurately reproducing from slight to strong expressions, and generalize to unseen motions.

This work develops on the generative model proposed in Otberdout~\etal~\cite{Otberdout_2022_CVPR}. Compared to this previous approach, the main novelties of this paper are:
\begin{itemize}
	\item we removed the constraint of starting the 4D sequence from a neutral face. Motion3DGAN was modified so that it can generate 4D transitions that switch between two generic expressions;
	\item we defined a strategy to augment the dataset of 4D expressions with interpolated, complex expressions;
	\item we expanded the experimental validation to three additional datasets, characterized by totally different expressions, identities and mesh topology; 
	\item we experimented more difficult scenarios, such as speech transfer and cross-dataset 3D reconstruction.
\end{itemize}

\section{Proposed Method}\label{sect:proposed-method}
\begin{figure*}[!ht]
\centering 
\includegraphics[width=1\linewidth]{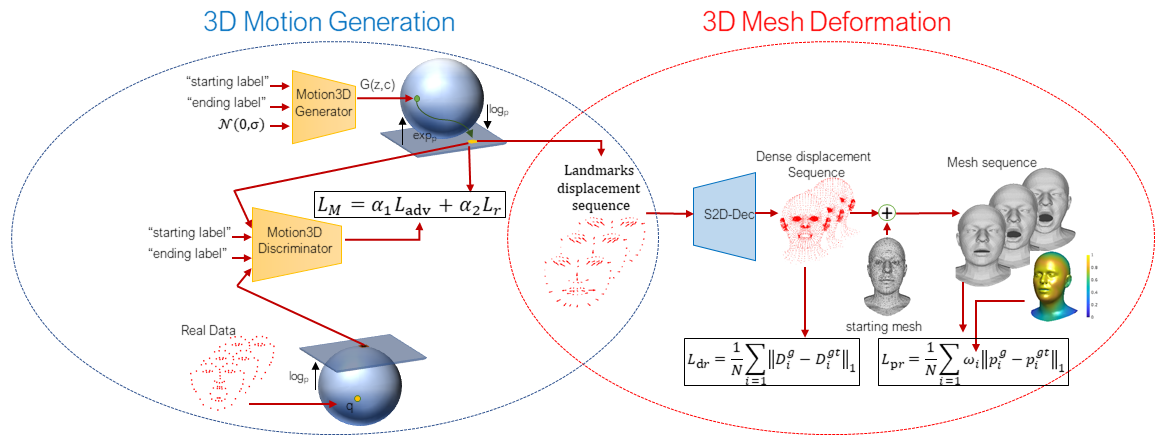}
\caption{\textbf{Overview of our framework}: Motion3DGAN generates the motion $q(t)$ of 3D landmarks corresponding to an expression label from a noise vector $z$. The module is trained guided by a reconstruction loss $L_r$ and adversarial loss $L_{adv}$. The motion $q(t)$ is converted to a sequence of landmark displacements $d_i$, which are fed to S2D-Dec. From each $d_i$, the decoder generates a dense displacement $D_i^g$. A neutral mesh is then summed to the dense displacements to generate the expressive meshes $S^g$. S2D-Dec is trained under the guidance of a displacement loss $L_{dr}$ and our proposed weighted reconstruction loss $L_{pr}$.}
\label{fig:detailed-approach}
\end{figure*}

Our approach consists of two specialized networks as summarized in Figure~\ref{fig:detailed-approach}. 
Motion3DGAN accounts for the temporal dynamics and generates the motion of a sparse set of 3D landmarks from noise. The generated motion represents a transition between two expressions defined by two labels, one for the start, \eg, neutral, happy, and the other for the ending configuration. The motion is then converted as a per-frame landmarks displacement. These displacements are then fed to a decoder network (S2D-Dec) that constructs the dense point-cloud displacements from the sparse displacements given by the landmarks. These dense displacements are finally added to a generic 3D face to generate a sequence of 3D faces corresponding to the specified transition from the starting expression to the ending one. In the following, we separately describe the two networks. 

\subsection{Generating 4D Expressions: Motion3DGAN}
Facial landmarks were shown to well encode the temporal evolution of facial expressions~\cite{kacem2017novel, OtberdoutPAMI2020}.  
Motivated by this fact, we generate the facial expression dynamics based on the motion of 3D facial landmarks. Given a set of $k$ 3D landmarks, $Z(t) = {(x_i(t),y_i(t),z_i(t))}_{i=1}^{k}$, with $Z(0)$ being the starting configuration, their motion can be seen as a trajectory in $\mathbb{R}^{k \times 3}$, and can be formulated as a parameterized curve in $\mathbb{R}^{k \times 3}$ space. 
Let $\alpha: I = [0,1] \rightarrow \mathbb{R}^{ k \times 3}$ represent the parameterized curve, where each $\alpha(t) \in \mathbb{R}^{k \times 3}$. For the purpose of modeling and studying our curves, we adopt the Square-Root Velocity Function (SRVF) proposed in~\cite{SrivastavaKJJ11}. The SRVF $q(t): I \rightarrow \mathbb{R}^{k \times 3}$ is defined by:
\begin{equation}
q(t)= \begin{cases}
\frac{\dot{\alpha}(t)}{\sqrt{\|\dot{\alpha}(t)\|_{}}}, & \text { if }\|\dot{\alpha}(t)\| \neq 0 \\ 0, & \text { if }\|\dot{\alpha}(t)\|=0 ,
\end{cases}
\label{curve_2_q}
\end{equation}

\noindent where, $\|\cdot\|$ is the Euclidean 2-norm in $\mathbb{R}^n$. This function proved effective for tasks such as human action recognition~\cite{devanne20143} or 3D face recognition~\cite{drira20133d}. Similar to this work, Otberdout~\etal~\cite{OtberdoutPAMI2020} proposed to use the SRVF representation to model the temporal evolution of 2D facial landmarks, which makes it possible to learn the distribution of these points and generate the motions for new 2D facial expression. 
In this paper, we extend this idea by proposing the Motion3DGAN model, which generates the motion of 3D facial landmarks. Differently from~\cite{OtberdoutPAMI2020}, where the dynamic expression is assumed to start from a neutral configuration, here we remove this constraint and train Motion3DGAN to generate motions corresponding to a transition between two expressions. The motion is represented using the SRVF encoding in~\eqref{curve_2_q}. Following~\cite{OtberdoutPAMI2020}, we remove the scale variability of the resulting motions by restricting curves $\alpha$ to length 1. As a result, we transform the motion of 3D facial landmarks to points on a Hilbert sphere of radius 1, $\mathcal{C}=\{q:[0,1] \rightarrow \mathbb{R}^{k \times 3}, \|q\|^2=1\}$. The geometry of sphere is well-understood and can be exploited.

To learn the distribution of the SRVF representations, we propose Motion3DGAN as an extension of MotionGAN~\cite{OtberdoutPAMI2020}, a conditional version of the Wasserstein GAN for manifold-valued data~\cite{HuangAAAI19}. It maps a random vector $z$ to a point on the Hilbert sphere $\mathcal{C}$ conditioned on an input labels pair $c = (start, end)$. Motion3DGAN is composed of two networks trained adversarially: a generator $G$ that learns the distribution of the 3D landmark motions, and a discriminator $D$ that distinguishes between real and generated 3D landmark motions. Motion3DGAN is trained by a weighted sum of an adversarial loss $L_{adv}$ and a reconstruction loss $L_{r}$ such that $L_{M}= \alpha_1 L_{adv} + \alpha_2 L_{r}$. 

The adversarial loss is $L_{adv}$ formulated as:
\begin{equation}
\begin{array}{rl}
L_{adv}= & {\mathbb{E}_{q \sim \mathbb{P}_{q}}\left[D\left(\log _{p}(q),c\right)\right]} 
\\ 
{ } & {-\mathbb{E}_{z \sim \mathbb{P}_{z}}\left[D\left(\log _{p}\left(\exp _{p}(G(z, c))\right)\right)\right]}
\\ 
{ } & {+\lambda E_{\hat{q} \sim \mathbb{P}_{\hat{q}}}\left[\left(\left\|\nabla_{\hat{q}} D(\hat{q})\right\|_{2}-1\right)^{2}\right]}.
\end{array}
\label{eq:LossWassersteinGAN}
\end{equation}

\noindent
In the above equation, the exponential map, $\exp_u(.)$: $T_{u}(C) \mapsto \mathcal{C}$ and its inverse, \ie, the logarithm map $\log _{u}(q)$: $\mathcal{C} \mapsto T_{u}(C)$ are used to map the SRVF data forth and back to a tangent space $T_{u}$ defined at a particular point $p$ of $\mathcal{C}$. They are computed as follows:  
\begin{equation}
\exp_{u}(s) = cos(\|s\|)u + sin(\|s\|)\frac{s}{\|s\|},
\label{eq:exponential}
\end{equation}

%
\begin{equation}
\log _{u}(q) = \frac{d_{\mathcal{C}}(q,u)}{sin(d_{\mathcal{C}}(q,r))} (q - cos(d_{\mathcal{C}}(q,u))u), 
\label{eq:logarithm}
\end{equation}

\noindent
where $d_{\mathcal{C}}(q,p)=\cos^{-1}(\langle q,p \rangle)$ is the geodesic distance between $q$ and $p$ in $\mathcal{C}$. 
In~\eqref{eq:LossWassersteinGAN}, $q \sim \mathbb{P}_{q}$ is an SRVF sample from the training set, $c$ is the expression labels pair (\eg, mouth open-eyebrow, bareteeth-mouth up) that is concatenated to a random noise $z \sim \mathbb{P}_{z}$. The last term of the adversarial loss represents the gradient penalty of the Wasserstein GAN~\cite{gulrajani2017improved}. 
Specifically, $\hat{q} \sim \mathbb{P}_{\hat{q}}$ is a random point sampled uniformly along straight lines between pairs of points sampled from $\mathbb{P}_{q}$ and the generated distribution $\mathbb{P}_{g}$:
\begin{equation}
\hat{q} = (1 - \tau) \log_p(q) + \tau\log_p(\exp_p(G(z, c))), 
\end{equation}

\noindent
where $0 \leq \tau \leq 1$, and $\nabla_{\hat{q}} D(\hat{q})$ is the gradient \wrt~$\hat{q}$. 

Finally, the reconstruction loss is defined as:
\begin{equation}
\label{eq:T_reconstruction_loss}
L_{r} = \|{ \log_{p}(\exp_p({G(z,c)})) - \log_p(q)} \|_1,
\end{equation}

\noindent
where $\|.\|_1$, represents the $L_1$-norm, and $q$ is the ground truth SRVF corresponding to the condition $c$. The generator and discriminator architectures are similar to~\cite{OtberdoutPAMI2020}.

The SRVF representation is reversible, which makes it possible to recover the curve $\alpha(t)$ from a new generated SRVF $q(t)$ by,
\begin{equation}
\label{eq:curve}
\alpha(t) = \int_0^t \norm{q(s)}q(s)ds + \alpha(0),
\end{equation}

\noindent 
where $\alpha(0)$ represents the initial landmark configuration $Z(0)$. Using this equation, we can apply the generated motion to \textit{any} landmark configuration, making it robust to identity changes.

\subsection{From Sparse to Dense 3D Expressions: S2D-Dec}
Our final goal is to animate the starting mesh $S^{n}$ to obtain a novel 3D face $S^{g}$ reproducing some expression, yet maintaining the identity structure of $S^{n}$. 
Given this, we point at generating the displacements of the mesh vertices from the sparse displacements of the landmarks to animate $S^{n}$. In the following, we assume all the meshes have a fixed topology, and are in full point-to-point correspondence. 

Let $\mathcal{L} = \left\{\left(S^{n}_1, S^{gt}_1, Z^n_1, Z^{gt}_1\right), \ldots,\left(S^{n}_{m}, S^{gt}_m, Z^{gt}_m, Z^n_m\right)\right\}$ be the training set, where $S^{n}_i = (p^n_1, \ldots, p^n_N) \in \mathbb{R}^{N \times 3}$ is a neutral 3D face, $S^{gt}_i = (p^{gt}_1, \ldots, p^{gt}_N) \in \mathbb{R}^{N \times 3}$ is a 3D expressive face, $Z^n_i \in \mathbb{R}^{k \times 3}$ and $Z^{gt}_i \in \mathbb{R}^{k \times 3}$ are the 3D landmarks corresponding to $S^{n}_i$ and $S^{gt}_i$, respectively. 
We transform this set to a training set of sparse and dense displacements, $\mathcal{L} = \left\{\left(D_1, d_1\right), \ldots, \left(D_{m}, d_m\right)\right\}$ such that, $D_i = S^{gt}_i - S^{n}_i$ and $d_i=Z^{gt}_i - Z^{n}_i$. Our goal here is to find a mapping $h: \mathbb{R}^{k \times 3} \rightarrow \mathbb{R}^{N \times 3}$ such that $D_i \approx h\left(d_i\right)$.
We designed the function $h$ as a decoder network (S2D-Dec), where the mapping is between a sparse displacement of a set of landmarks and the dense displacement of the entire mesh points. Finally, in order to obtain the expressive mesh, the dense displacement map is summed to a 3D face in neutral expression, \ie, $S^{e}_i = S^{n}_i + D_i$. 
The S2D-Dec network is based on the spiral operator proposed in~\cite{Bouritsas:ICCV2019}. Our architecture includes five spiral convolution layers, each one followed by an up-sampling layer. 
More details on the architecture can be found in the supplementary material.

In order to train this network, we propose to use two different losses, one acting directly on the displacements and the other controlling the generated mesh. The reconstruction loss of the dense displacements is given by,
\begin{equation}
\label{eq:displacement_loss}
L_{dr} = \frac{1}{N}\sum_{i=1}^{N}  \left \| D^{g}_i - D^{gt}_i \right \|_1,
\end{equation}

\noindent
where $D^{g}$ and $D^{gt}$ are the generated and the ground truth dense displacements, respectively. To further improve the reconstruction accuracy, we add a loss that minimizes the error between $S^{g}$ and the ground truth expressive mesh $S^{gt}$. We observed that vertices close to the landmarks are subject to stronger deformations. Other regions like the forehead, instead, are relatively stable. To give more importance to those regions, we defined a weighted version of the $L1$ loss:
\begin{equation}
\label{eq:L1_weighted}
L_{pr}= \frac{1}{N}\sum_{i=1}^{N} w_i \cdot \left \| p^{g}_i - p^{gt}_i \right \|_1 .
\end{equation}

\noindent
We defined the weights as the inverse of the Euclidean distance of each vertex $p_i$ in the mesh from its closest landmark $Z_j$, \ie $w_i = \frac{1}{\min d(p_i, Z_j)}, \; \forall j$. This provides a coarse indication of how much each $p_i$ contributes to the expression generation. Since the mesh topology is fixed, we can pre-compute the weights $w_i$ and re-use them for each sample. Weights are then re-scaled so that they lie in $[0, 1]$. Vertices corresponding to the landmarks, \ie, $p_i = Z_j$ for some $j$, are hence assigned the maximum weight. 
We will show this strategy provides a significant improvement with respect to the standard $L1$ loss. The total loss used to train the S2D-Dec is given by $L_{S2D}=\beta_1. L_{dr} + \beta_2. L_{pr}$.

\section{Experiments}\label{sect:results}
We validated the proposed method in a broad set of experiments on five publicly available benchmark datasets.

\noindent
\textbf{CoMA dataset}~\cite{COMA:ECCV18}: It is a common benchmark employed in other studies~\cite{Bouritsas:ICCV2019, COMA:ECCV18}. It includes $12$ subjects, each one performing $12$ extreme and asymmetric expressions. Each expression comes as a sequence of meshes $S \in \mathbb{R}^{N \times 3}$ ($140$ meshes on average), with $N = 5,023$ vertices. Sequences start from a neutral state, reach a peak of the expression, and then get back to a neutral state.\\
\textbf{D3DFACS dataset}~\cite{cosker2011facs}. We used the registered version of this dataset~\cite{li2017learning}, which has the same topology of CoMA. It contains $10$ subjects, each one performing a different number of facial expressions. In contrast to CoMA, this dataset is labeled with the activated action units of the performed facial expression. It is worthy to note that the expressions of D3DFACS are highly different from those in CoMA. \\
\textbf{Florence 4D Facial Expression dataset (Florence 4D)}~\cite{Florence:fg2022}. This dataset consists of 10,710 synthetic sequences of 3D faces with different facial expressions from which we selected 1,222 sequences corresponding to the 7 standard facial expressions: angry, disgust, fear, happy, sad and surprise. The sequences correspond to 155 subjects including 117 females and 38 males. Each sequence is composed of $60$ frames showing an expression that evolves from neutral face to reach the peak and then get back to the neutral state. The meshes are in full correspondence with the Flame template. The dataset includes synthetic identities based on the DAZ Studio’s Genesis 8 Female~\cite{daz3d} as well as CoMa identities and real scans from the Florence 2D/3D dataset~\cite{Bagdanov:2011}. Expressions were generated with the DAZ Studio software~\cite{daz3d}.\\
\textbf{VOCASET}. This dataset provides 480 speech sequences of 3D face scans belonging to the 12 identities of CoMA dataset. The faces are in full correspondence and aligned to the Flame template. \\
\textbf{BU-3DFE}. This dataset contains scans of 44 females and 56 males, ranging from 18 to 70 years old, acquired in neutral plus the prototypical six expressions. Each of the six expressions is acquired at four levels of intensity. Those, however, are not in full, point-to-point correspondence. For the sake of this work, we employed the registered version as described in~\cite{ferrari2015dictionary}, which includes $1,779$ meshes, each mesh having $N = 6704$ vertices. We underline that we chose this particular dataset in addition to the previous ones to show our S2D-Dec can effectively handle different mesh topology, and is robust to possible noise as can result from a dense registration process. 

\begin{figure}[!t]
\centering
\includegraphics[width=0.9\linewidth]{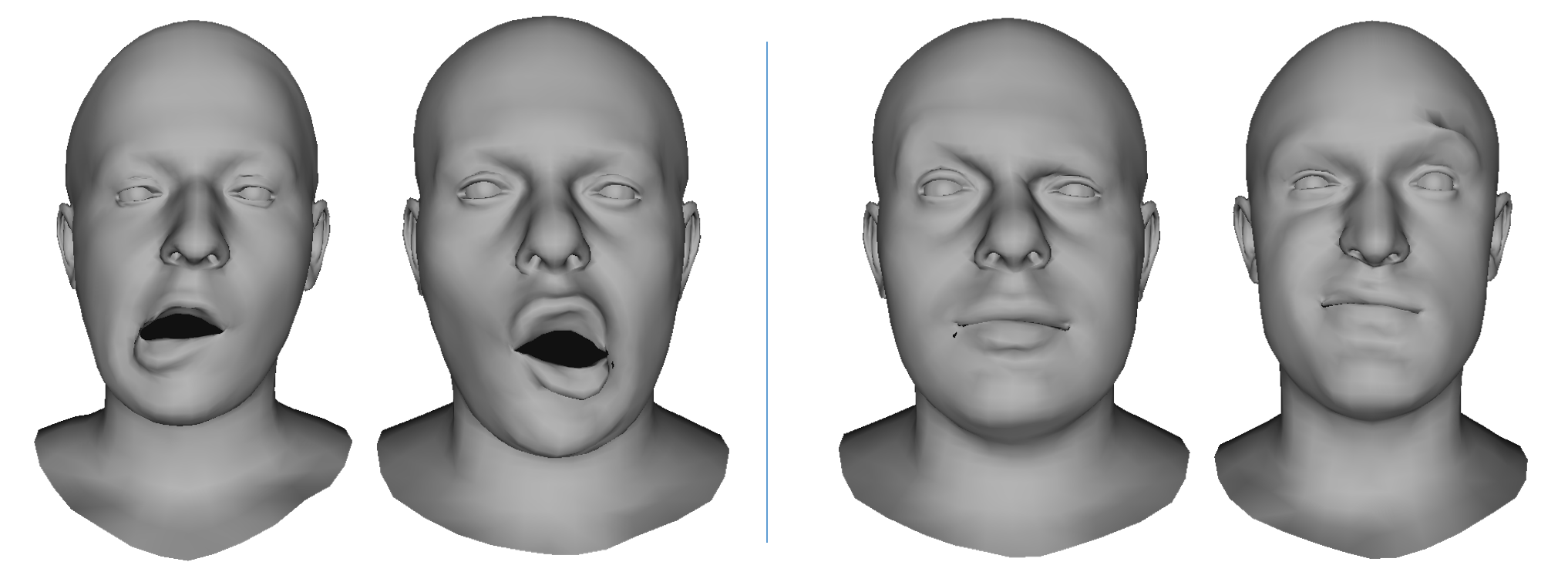}
\caption{Samples from the CoMA dataset: Mouth-Side (left), Eyebrows (right). For the same expressions, the samples differ significantly.}
\label{fig:divExp}
\end{figure}

\subsection{Training Details}\label{sec:Train_details}
In order to keep Motion3DGAN and S2D-Dec decoupled, they are trained separately. 

\smallskip

\noindent
\textbf{Motion3DGAN}: We used CoMA to train Motion3DGAN, since this dataset contains 4D sequences labeled with facial expression classes. However, in order to train Motion3DGAN to generate transitions that shift from one expression to another, the CoMA dataset in its original form is not suitable, since it only includes \textit{neutral-peak-neutral} sequences, whereas we also need \textit{peak-peak} transitions. So, we defined a solution to expand the dataset to include such peak-peak transitions.

In particular, we processed the dataset as follows: first, we manually divided the existing sequences into sub-sequences of $30$ frames/meshes. They are separated in neutral to peak of the expression (neutral-peak), and vice versa (peak-neutral). These are then encoded as points $q$ on a hyper-sphere using the SRVF representation in~\eqref{curve_2_q}. Note that, for neutral-peak sequences, the initial landmark configuration required to compute the SRVF motion for a subject $j$ is the neutral one, \ie, $Z_j(0) = Z_j^n$, while for peak-neutral motions, the initial landmark configuration needs to be that of the peak frame of the expression, \ie, $Z_j(0) = Z_j^{e}$. In this way, we can easily interpolate two sequences: given two points on the sphere $q_1$ and $q_2$, representing two expression motions $e_1, e_2$ (either neutral-peak or peak-neutral), the geodesic path $\psi(\tau)$ between them is given by: 
\begin{equation}
    \psi(\tau)=\frac{1}{sin(\theta)}sin((1-\tau)\theta) q_1 + sin(\tau\theta)q_2, 
\end{equation}%
where, $\theta=d_{\mathcal{C}}\left(q_{1}, q_{2}\right)=\cos ^{-1}\left(\left\langle q_{1}, q_{2}\right\rangle\right)$, and $\tau \in [0,1]$. This path determines all the points $q_i$ existing between $q_1$ and $q_2$, each of them corresponding to a (interpolated) landmarks motion. We do so for all the 12 subjects and their 12 expressions.

\begin{figure}[!t]
\centering
\includegraphics[width=0.99\linewidth]{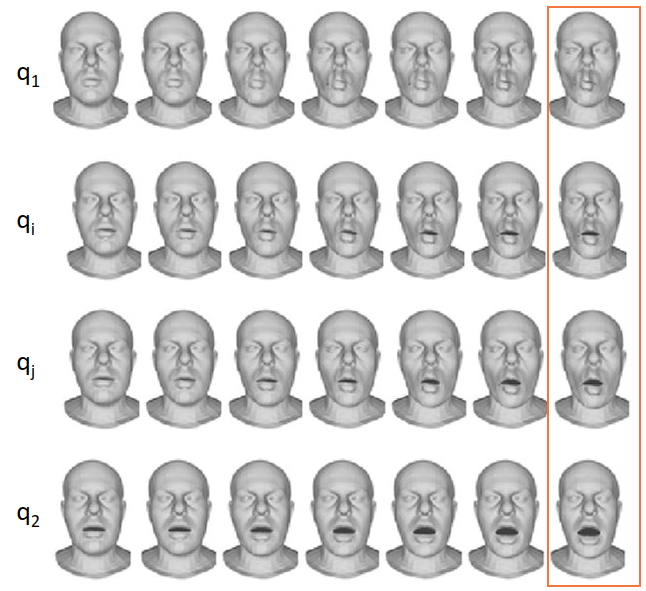}
\caption{Example of the generation of a peak-peak transition across two given expression sequences. Given two expression motions, $q_1$ and $q_2$, we generate 30 interpolated sequences by sampling the geodesic path between their SRVF representation. The transition between the two peaks is obtained by converting back the motion to landmarks, and concatenating the last frame of each sequence (red rectangle). Meshes generated with S2D-Dec are visualized in place of landmarks for clarity.}
\label{fig:interp}
\end{figure}

We then estimate the transition between two expression peaks. The idea is that each interpolated motion $q_i$ ends to an expression peak, which is a linear combination of the peaks of two generic motions $q_1,q_2$. Thus, if we generate a certain number of motions $q_i$ at different interpolation steps, convert them back to landmarks using~\eqref{eq:curve}, and keep the last frame of each sequence $q_i$, then their concatenation provides us a transition between the peaks of $q_1$ and $q_2$. In order to collect compatible sequences, we generated $30$ interpolated motions $q_i$ for each possible pair, so that the peak-peak transition is of the same length of the original ones, \ie, 30 frames. This process is depicted in Figure~\ref{fig:interp}. This sequence is then converted to a point in the SRVF space for training Motion3DGAN. Doing so for all possible pairs resulted in approximately $300K$ samples in total. We chose to perform interpolation directly in the hypersphere induced by SRVF as it leads to cleaner results than interpolating directly in the 3D space~\cite{pierson2022riemannian}. However, we cannot use all of them. In fact, different subjects might perform the same expression in a significantly different way, as shown in Figure~\ref{fig:divExp}. To obtain a correct peak-peak transition though, the initial landmark configuration $Z(0)$ must be coherent with the estimated motion, which can only be guaranteed by using identity-specific landmarks.
In order to maintain full independence from the identity, we instead computed the average landmark configuration of the expression peaks across subjects, for each expression. We then used these prototypes to select the most similar ones among the interpolated transitions, obtaining a total of $6,740$ interpolated motions. One sample for each transition is used as test set. We used all the others for training as we generate from a random noise at test time.

\begin{table*}[!t]
\centering
\caption{Reconstruction error (mm) on expression-independent (left) and identity-independent (right) splits: comparison with PCA-$k$ 3DMM ($k$ components), DL-3DMM (220 dictionary atoms), and Neural3DMM.}
\label{tab:comparison-ExprSplit}
\small
\begin{tabular}{@{}l@{}cccc@{}|cccc@{}}
\toprule
& \multicolumn{3}{c}{Expression Split} & \multicolumn{3}{c}{Identity Split}\\
\midrule
Method & CoMA & D3DFACS & BU-3DFE & Florence 4D & CoMA & D3DFACS & BU-3DFE & Florence 4D\\
\midrule
PCA-220 & $0.76 \pm 0.73$  &$0.42 \pm 0.44$ & $2.00\pm 1.67$& $0.70\pm 0.81$ & $0.80 \pm 0.73$ & $0.56 \pm 0.56$ & $\mathbf{2.10\pm1.74}$ & $0.16\pm 0.17$ \\
PCA-38 & $0.90 \pm 0.84$ &$0.44\pm0.45 $ &$2.55\pm 1.72$ & $0.87\pm 1.05$ & $0.93 \pm 0.82$ & $0.58 \pm 0.56$ & $2.61\pm 1.78 $& $0.18\pm 0.20$\\
DL3DMM~\cite{ferrari2017dictionary} & $0,86 \pm 0,80$  & $0.73\pm 1.15$ & $2.09 \pm 1.58$ & $0.83\pm 1.03$ & $0.89 \pm 0.79$ & $1.15 \pm 1.50$ & $2.22 \pm 1.69$ & $0.17\pm 0.18$\\
Neural~\cite{Bouritsas:ICCV2019} & $0.75 \pm 0.85$ & $0.59 \pm 0.86$ &$3.16\pm 2.12$& $1.45\pm 1.43$ & $3.74 \pm 2.34$ & $2.09\pm 1.37$ & $3.85\pm2.32$ &$1.41\pm 1.09$ \\
\midrule
{\bf Ours} & $\mathbf{0.52 \pm 0.59}$ & $\mathbf{0.28\pm 0.31}$ &$\mathbf{1.97\pm 1.52}$& $\mathbf{0.57\pm 1.24}$ & $\mathbf{0.55 \pm 0.62}$ & $\mathbf{0.27 \pm 0.30}$ & $2.34\pm1.83$ & $\mathbf{0.10\pm 0.08}$\\
\bottomrule
\end{tabular}
\end{table*}

To train the model, we encoded the motions of $k=68$ landmarks in the SRVF representation. The landmarks were first centered and normalized to unit norm. To encode the starting-ending labels pair, each of the $13$ expression (including neutral) was first encoded as a one-hot vector; the labels pairs is formed by concatenating two expression labels. Finally, they are further concatenated with a random noise vector of size $128$. 

\smallskip

\noindent
\textbf{S2D-Dec:} To comprehensively evaluate the capability of S2D-Dec of generalizing to either unseen identities or expressions, we performed subject-independent and expression-independent cross-validation experiments. For the \textit{subject-independent} experiment, we used a $4$-fold cross-validation protocol for CoMA, training on $9$ and testing on $3$ identities in each fold. On D3DFACS, we used the last $7$ identities for training and the remaining $3$ as test set. For the Florence 4D dataset, we used the last 30 females and 10 males as test set, and trained the model on the remaining subjects. Finally, the test set of the BU-3DFE includes the first 10 subjects (5 females and 5 males). For the \textit{expression-independent} splitting, we used a $4$-fold  cross-validation protocol for CoMA, training on $9$ and testing on $3$ expressions in each fold. For D3DFACS, given the different number of expressions per subject, the first $11$ expressions were used for testing and the remaining expressions were used for training. For Florence 4D, the first two expressions of each subject were used as test set. Finally, the test set of BU-3DFE includes the two expressions Angry and Disgust, while the remeaining expressions were used for training.

\smallskip

We trained both Motion3DGAN and S2D-Dec using the Adam optimizer, with learning rate of $0.0001$ and $0.001$ and mini-batches of size $128$ and $16$, respectively. Motion3DGAN was trained for $8,000$ epochs, while $300$ epochs were adopted for S2D-Dec. The hyper-parameters of the Motion3DGAN and S2D-Dec losses were set empirically to $\alpha_1 = 1$, $\alpha_2 = 10$, $\beta_1= 1$ and $\beta_2=0.1$. We chose the mean SRVF of the CoMA data as a reference point $p$, where we defined the tangent space of $\mathcal{C}$. 


\subsection{3D Expression Generation: S2D-Dec}
For evaluation, we set up a baseline by first comparing against standard 3DMM-based fitting methods. Similar to previous works~\cite{ferrari2017dictionary, FLAME:SiggraphAsia2017}, we fit $S^n$
to the set of target landmarks $Z^{e}$ using the 3DMM components. Since the deformation is guided by the landmarks, we first need to select a corresponding set from $S^n$ to be matched with $Z^{e}$. Given the fixed topology of the 3D faces, we can retrieve the landmark coordinates by indexing into the mesh, \ie, $Z^n = S^n(\mathbf{I}_z)$, where $\mathbf{I}_z \in \mathbb{N}^{n}$ are the indices of the vertices that correspond to the landmarks. We then find the optimal deformation coefficients that minimize the Euclidean error between the target landmarks $Z^{e}$ and the neutral ones $Z^{n}$, and use the coefficients to deform $S^n$. 
In the literature, 
several 3DMM variants have been proposed. We experimented the standard PCA-based 3DMM and the DL-3DMM in~\cite{ferrari2017dictionary}. 
For fair comparison, we built the two 3DMMs using a number of deformation components comparable to the size of the S2D-Dec input, \ie, $68 \times 3 = 204$. For PCA, we used either $38$ components ($99\%$ of the variance) and $220$, while for DL-3DMM we used $220$ dictionary atoms.

With the goal of comparing against other deep models, we also considered the Neural3DMM~\cite{Bouritsas:ICCV2019}. It is a mesh auto-encoder tailored for learning a non-linear latent space of face variations and reconstructing the input 3D faces. In order to compare it with our model, we modified the architecture and trained the model to generate an expressive mesh $S^g$ given its neutral counterpart as input. To do so, we concatenated the landmarks displacement (of size $204$) to the latent vector (of size $16$) and trained the network towards minimizing the same $L_{pr}$ loss used in our model. 
We used the same data to train all the models for consistency. However, since we exclude identity reconstruction in our problem, it could be argued that multi-linear 3DMMs, where identity and expressions are handled by two different models, should be used. We also experimented by building expression-specific 3DMMs, obtained by subtracting the neutral scan of each subject from their expressive counterparts instead of using the overall data mean. However, this not resulted in any noticeable improvement.
Finally, we also identified the FLAME model~\cite{FLAME:SiggraphAsia2017}. Unfortunately, the training code of FLAME is not available, while using the model pre-trained on external data would result in an unfair comparison. 

The mean per-vertex Euclidean error between the generated meshes and their ground truth is used as standard performance measure, as in the majority of works~\cite{Bouritsas:ICCV2019, COMA:ECCV18, ferrari2021sparse, PotamiasECCV2020}. Note that we exclude the Motion3DGAN model here as we do not have the corresponding ground-truth for the generated landmarks (they are generated from noise). Instead, we make use of the ground truth motion of the landmarks.

\begin{figure}[!t]
\centering
\includegraphics[width=\linewidth]{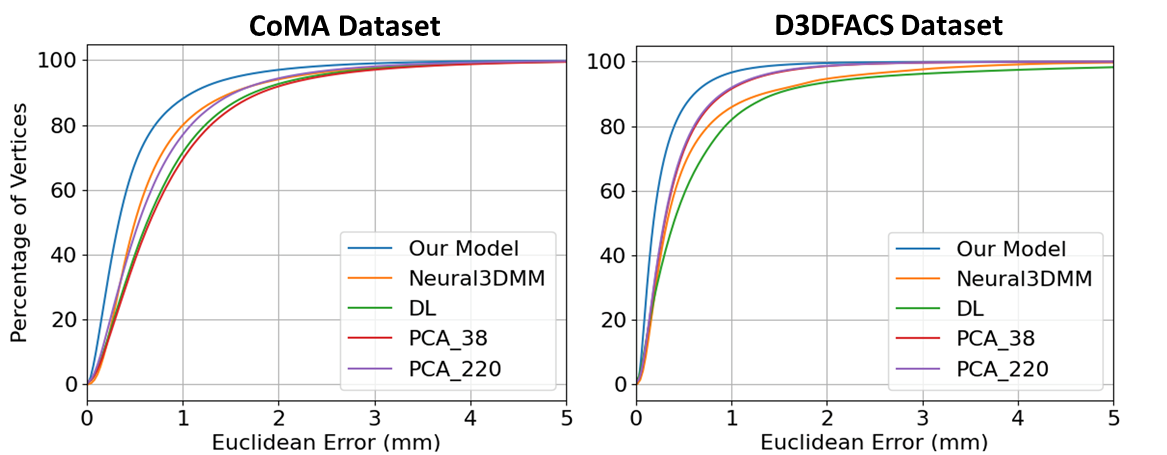}\\
\includegraphics[width=\linewidth]{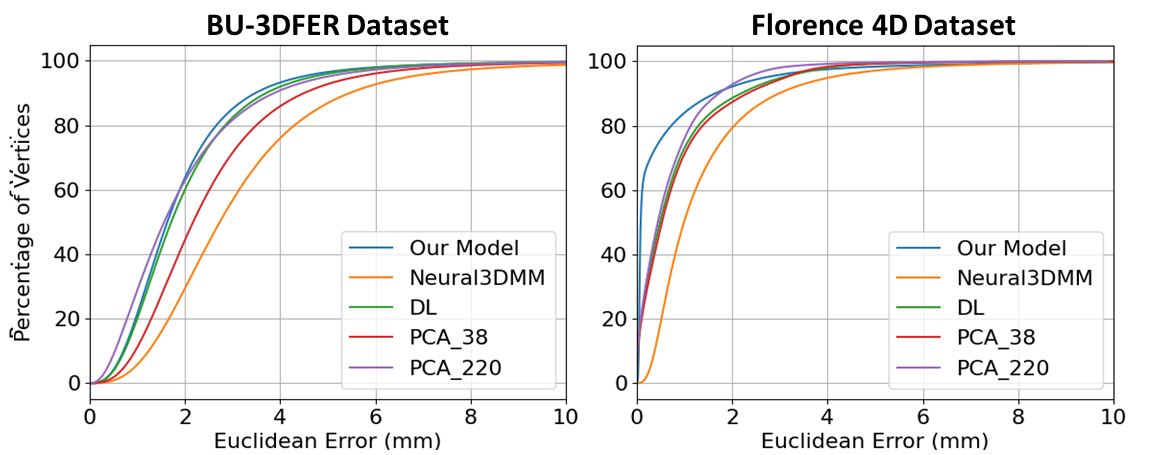}
\caption{Cumulative per-vertex error between PCA-based 3DMM models, DL-3DMM, Neural3DMM, and S2D-Dec, using expression-independent cross-validation on four datasets.}
\label{fig:cum-err-exprind}
\end{figure}

\subsubsection{Comparison with Other Approaches}
Table~\ref{tab:comparison-ExprSplit} shows a clear superiority of S2D-Dec over state-of-the-art methods for both the protocols and datasets, proving its ability to generate accurate expressive meshes close to the ground truth in both the case of unseen identities or expressions. In Figure~\ref{fig:cum-err-exprind}, the cumulative per-vertex error distribution on the expression-independent splitting further highlights the precision of our approach, which can reconstruct 90\%-98\% of the vertices with an error lower than $1mm$. While other fitting-based methods retain satisfactory precision in both the protocols, we note that the performance of Neural3DMM~\cite{Bouritsas:ICCV2019} significantly drops when unseen identities are considered. This outcome is consistent to that reported in~\cite{ferrari2021sparse}, in which the low generalization ability of these models is highlighted. Overall, our solution embraces the advantages of both approaches, being as general as fitting solutions yet more accurate. The only case where our method performs slightly worse is for the BU-3DFE dataset. Here, meshes are obtained through a dense registration, which is an error-prone process mostly for expressive scans. So, training data is likely affected by noise. However, results show S2D-Dec is quite robust.

Figure~\ref{fig:cum-err} shows some qualitative examples by reporting error heatmaps in comparison with PCA, DL-3DMM~\cite{ferrari2017dictionary} and Neural3DMM~\cite{Bouritsas:ICCV2019} for the identity-independent splitting. The ability of our model as well as PCA and DL-3DMM to preserve the identity of the ground truth comes out clearly, in accordance with the results in Table~\ref{tab:comparison-ExprSplit}. By contrast, Neural3DMM shows high error even for neutral faces, which proves its inability to generalize to the identity of unseen identities. Indeed, differently from the other methods, Neural3DMM encodes the neutral face in a latent space and predicts the 3D coordinates of the points directly. This evidences the efficacy of our S2D-Dec that learns per-point displacements instead of point coordinates. 

\begin{figure}[!t]
\centering
\includegraphics[width=0.9\linewidth]{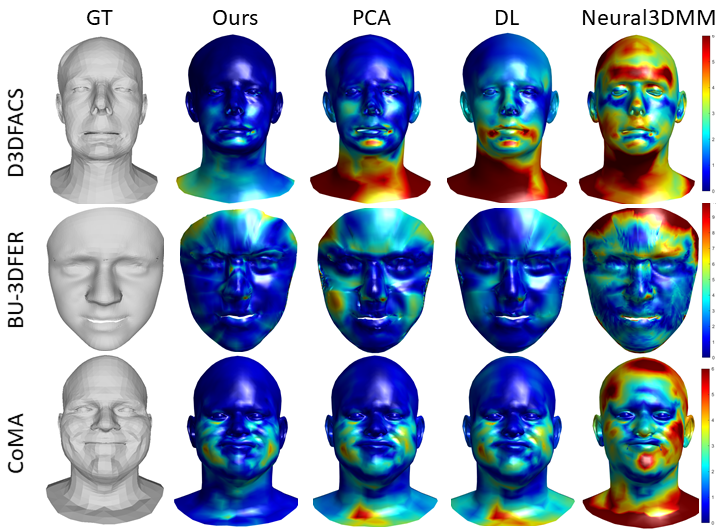}
\caption{Mesh reconstruction error (red=high, blue=low) of our model and other methods. Examples are given for three different datasets.}
\label{fig:cum-err}
\end{figure}

\subsubsection{Transfer of Speech-Related Facial Movements}\label{subsubsec:speech}
By using landmarks, our S2D-Dec can transfer facial expressions or speech between identities. This is done by extracting the sequence of landmarks from the source face, encoding their motion as an SRVF representation, transferring this motion to the neutral landmarks of the target face and using S2D-Dec to get the target identity following the motion of the first one. To demonstrate the high generalization ability of our proposed approach for different expressions, we evaluate it on VOCASET for speech transfer. This is done by transferring the speech-related movements from the first identity of VOCASET to the other 11 identities.

We report the reconstruction error between the obtained meshes after speech transfer and their ground truth counterparts. Note that the lengths of the ground truth and the obtained sequences are slightly different, thus we considered the error for each frame as the minimum error in a sliding windows of $20$ frames centered on the given frame. In this experiment, we only considered the first five sentences of VOCASET that are shared between all identities. We highlight that the model used in this experiment was trained on the CoMA dataset that does not include such speech-related movements. That is possible given the full correspondence between CoMA and VOCASET data. Table~\ref{tab:VOCASET_results} shows the results of this experiment. The superiority of our approach is clearly evidenced over other state-of-the-art solutions, which proves the high generalization ability of our method to animate 3D faces with completely different facial expressions from those seen during the training. In addition, these results demonstrate that our method can be used not only with our generated facial motions, but we can also exploit external motions that are completely different from our generated ones.

\begin{table}[!t]
\centering
\caption{Speech transfer reconstruction error on VOCASET}
\label{tab:VOCASET_results}
\small
\begin{tabular}{lcccc}
\toprule
 - & PCA & DL3DMM & Neural & Ours (S2D)\\
\midrule
Rec. error (mm) & $2.90$ & $2.73$ & $3.49$ & \textbf{$1.48$} \\
\bottomrule
\end{tabular}
\end{table}

\subsubsection{Cross Datasets Evaluation} 
We report the error obtained for a cross-dataset evaluation on two different datasets; COMA and Florence 4D. 
The error is reported on all CoMA samples and the test set of  Florence 4D. In consistency with the previous results, Table~\ref{tab:cross-dataset} confirms the superiority of our model over other methods. We note that the mean errors obtained with the Florence 4D data are almost the same obtained with the expression split protocol in Table~\ref{tab:comparison-ExprSplit}. However, a higher error is reached on CoMA with all methods. 

\begin{table}[!t]
\centering
\caption{Reconstruction error (mm): cross dataset setting}
\label{tab:cross-dataset}
\begin{tabular}{l*{2}{c}}
\toprule
Method & \makecell{Train: Florence 4D \\ Test: CoMA} & \makecell{Train: CoMA \\ Test: Florence 4D} \\
\midrule
PCA 220 & $1.81 \pm 1.54$ & $0.64 \pm 0.89$ \\
DL & $2.09 \pm 1.78$ & $0.72 \pm 0.91$ \\
Ours & $1.50 \pm 1.84$ & $0.56 \pm 0.76$ \\
\bottomrule
\end{tabular}
\end{table}

\subsubsection{Ablation Study}
We report here an ablation study to highlight the contribution of each loss used to train S2D-Dec, with particular focus on our proposed weighted-$L1$ reconstruction loss. We conducted this study on the CoMA dataset using the first three identities as a testing set and training on the rest. This evaluation is based on the mean per-vertex error between the generated and the ground truth meshes. We evaluated three baselines, namely, $S1$, $S2$ and $S3$. For the first baseline ($S1$), we trained the model with the displacement reconstruction loss in~\eqref{eq:displacement_loss} only. In $S2$, we added the standard $L1$ loss to $S1$, which corresponds to our loss in~\eqref{eq:L1_weighted} without the landmark distance weights. To showcase the importance of weighting the contribution of each vertex, in $S3$ we added the landmark distance weights to the $L_{pr}$ loss. 
Results are shown in Table~\ref{tab:Ablation-study}, where the remarkable improvement of our proposed loss against the standard $L1$ turns out evidently. This is explained by the fact that assigning a greater weight to movable face parts allows the network to focus on regions that are subject to strong facial motions, ultimately resulting in realistic samples. 

\begin{table}[!t]
\centering
\caption{Ablation study on the reconstruction loss of S2D-Dec}
\label{tab:Ablation-study}
\small
\begin{tabular}{@{}lc@{}}
\toprule
Method & Error (mm) \\
\midrule
$S_1$ : $L_{dr}$ & $1.27 \pm 1.88$\\
$S_2$ : $S_1  + L_{pr}$ w/o distance weights & $0.92 \pm 1.33$ \\
$S_3$ :  $S_1 + L_{pr} $  & $0.50\pm 0.56$\\
\bottomrule
\end{tabular}
\end{table}

\subsection{4D Facial Expressions: Motion3DGan}
We validated the performance of Motion3DGAN in a broad set of experiments, both quantitative and qualitative. However, since Motion3DGAN generates samples from noise to encourage diversity, the generated landmarks and meshes change at each forward pass. Thus, we cannot directly compute the mean per-vertex error with respect to ground-truth shapes as done in~\cite{PotamiasECCV2020}. Comparing with other approaches is also not possible since no other method currently can generate dynamic transition sequences of arbitrary expressions. For a comprehensive analysis, we evaluated it in terms of \textit{(i)} specificity error, and \textit{(ii)} expression classification.

\smallskip

\noindent
\textbf{Specificity measure}: Following the standard practice for statistical generative shape models, we use the \textit{specificity} measure~\cite{davies2008statistical} to evaluate the quality of the generated samples. Given the very large number of possible start-end transitions (132 for the $12$ expressions of CoMA), we selected a subset of them for validation. In particular, for each expression, we randomly chose 3 possible ending expressions, obtaining a total of 39 transitions. For each transition, we generated $64$ samples (landmark sequences), for a total of $2,496$ samples, and computed the per-landmark average Euclidean distance with respect to the same transitions in the test data (as defined in Section~\ref{sec:Train_details}).
The average errors for all the cases are reported in Table~\ref{tab:specificity}. We first observe the error is stable and consistent across all the tested combinations. In addition, results show that transitions starting from the neutral expression score a lower distance. This is because the neutral expression is consistent across identities, while each subject performs facial expressions differently. In many cases, these can differ significantly; for example, some subjects of CoMA perform the ``Eyebrows'' expression by raising both of them, some others raise either the left or the right one (see Figure~\ref{fig:divExp}). Recalling~\eqref{eq:curve}, to obtain the landmarks from the generated motions, a reference landmark configuration for each expression needs to be chosen. Whereas for those starting from neutral this is not an issue, if the reference differs from that of the specific subject, the error might be larger even though the sequence is correct. To verify this, we performed a classification test, described in the next paragraph. 

\begin{figure}[!t]
\centering
\includegraphics[width=0.99\linewidth]{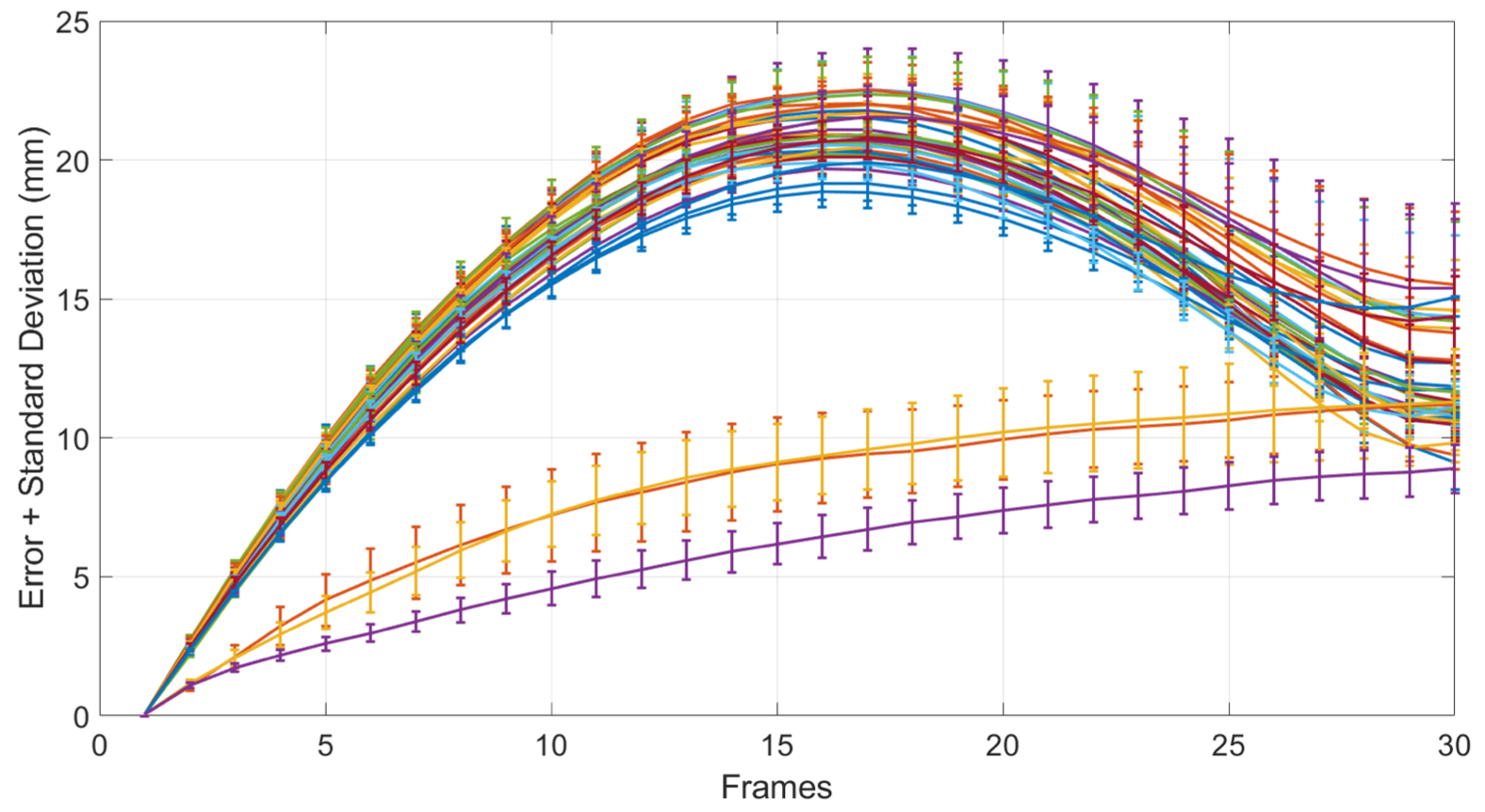}
\caption{Per-frame specificity error (mm) for all the 39 tested transitions. Legend is omitted for clarity. The three curves with lower error (purple, yellow, orange) refer to transitions starting from the neutral expression.}
\label{fig:per-frame-err}
\end{figure}

In Figure~\ref{fig:per-frame-err}, the per-frame specificity error is reported. It can be observed as, even though the three sequences starting from neutral obtain lower error on average, the error does not diverge. The higher increase in correspondence of the central frames (10-20) is again due to the very different and personal nature of facial expressions, which depends also on the velocity of performing it. The onset phase thus results more problematic, while at the peak of the expression (frame 30) the error tends to converge to a uniform value. However, the need of using an initial configuration of landmarks can be considered as a limitation of our approach; solving it would require generating also the starting configuration to ensure an even more pronounced diversity.

\smallskip

\noindent
\textbf{Expression classification}: We further evaluated the quality of the generated sequences implementing a classification solution, similar to~\cite{PotamiasECCV2020}. We trained a simple random forest classifier to recognize the $39$ transitions generated in the previous paragraph. We trained this classifier on the same sequences used to train Motion3DGAN. For testing, we used the same $2,496$ samples. Since the S2D-Dec could compensate minor generation errors, we directly used the generated landmarks to perform classification.

Results are reported in Table~\ref{tab:specificity}, separately for each transition. Overall, the generated sequences have a high classification accuracy. meaning that they accurately resemble real ones, even though some of them score a lower accuracy. Qualitatively, we verified this is likely caused by the similarity of some classes of expressions in the CoMA dataset. For example, mouth extreme qualitatively looks similar to a combination of mouth open and mouth down, just differing in intensity.

\begin{table}[!t]
\centering
\caption{Specificity error (mm) and classification accuracy for 39 transitions}
\label{tab:specificity}
\begin{adjustbox}{width=\columnwidth, center}
\begin{tabular}{@{}ll|c|c@{}}
\toprule
Start & End & Specificity (mm) & Classification \\
\midrule
\multirow{3}{*}{Bareteeth} & High-Smile & $15.1 \pm 0.77$ & $98\%$\\
                           & Lips-Back  & $15.3 \pm 0.52$ & $92\%$\\
                           & Lips-Up    & $14.4 \pm 0.72$ & $100\%$\\
\midrule
\multirow{3}{*}{Cheeks-In} & Mouth-Down     & $16.4 \pm 1.54$ & $80\%$\\
                           & Mouth-Extreme  & $14.1 \pm 0.41$ & $86\%$\\
                           & Mouth-Middle   & $16.3 \pm 0.13$ & $75\%$\\
\midrule
\multirow{3}{*}{Eyebrow}   & Mouth-Open     & $14.7 \pm 0.36$ & $100\%$\\
                           & Mouth-Side     & $15.6 \pm 0.72$ & $91\%$\\
                           & Mouth-Up       & $16.1 \pm 1.02$ & $86\%$\\
\midrule
\multirow{3}{*}{High-Smile}& Neutral     & $14.3 \pm 0.36$ & $100\%$\\
                           & Bareteeth   & $15.6 \pm 0.72$ & $100\%$\\
                           & Cheeks-In   & $16.1 \pm 1.02$ & $99\%$\\
\midrule
\multirow{3}{*}{Lips-Back} & Eyebrow     & $14.3 \pm 0.46$ & $79\%$\\
                           & High-Smile  & $14.4 \pm 0.59$ & $98\%$\\
                           & Lips-Up     & $13.8 \pm 0.59$ & $100\%$\\
\midrule
\multirow{3}{*}{Lips-Up}   & Mouth-Down     & $16.0 \pm 0.47$ & $100\%$\\
                           & Mouth-Extreme  & $14.0 \pm 1.09$ & $81\%$\\
                           & Lips-Back     & $14.4 \pm 0.45$ & $76\%$\\
\midrule
\multirow{3}{*}{Mouth-Down}  & Mouth-Middle & $16.4 \pm 1.51$ & $86\%$\\
                           & Mouth-Open     & $15.0 \pm 0.40$ & $95\%$\\
                           & Mouth-Side     & $15.6 \pm 0.61$ & $89\%$\\
\midrule
\multirow{3}{*}{Mouth-Extreme}  & Mouth-Up      & $15.9 \pm 0.47$ & $80\%$\\
                           & Neutral     & $14.0 \pm 1.20$ & $98\%$\\
                           & Bareteeth     & $14.5 \pm 0.01$ & $100\%$\\
\midrule
\multirow{3}{*}{Mouth-Middle}  & Cheeks-In  & $14.6 \pm 0.56$ & $100\%$\\
                           & Eyebrow        & $14.5 \pm 0.47$ & $75\%$\\
                           & High-Smile     & $14.6 \pm 0.48$ & $100\%$\\
\midrule
\multirow{3}{*}{Mouth-Open}  & Lips-Back    & $14.7 \pm 0.45$ & $100\%$\\
                           & Lips-Up        & $13.6 \pm 0.54$ & $100\%$\\
                           & Mouth-Down     & $16.0 \pm 1.06$ & $82\%$\\
\midrule
\multirow{3}{*}{Mouth-Side}  & Mouth-Extreme    & $15.2 \pm 0.43$ & $97\%$\\
                           & Mouth-Middle       & $15.9 \pm 1.28$ & $78\%$\\
                           & Mouth-Open         & $15.0 \pm 0.43$ & $100\%$\\
\midrule
\multirow{3}{*}{Mouth-Up}  & Mouth-Side      & $15.1 \pm 0.52$ & $88\%$\\
                           & Neutral     & $13.4 \pm 0.75$ & $100\%$\\
                           & Bareteeth     & $14.1 \pm 0.08$ & $100\%$\\
\midrule
\multirow{3}{*}{Neutral}  & Bareteeth      & $7.9 \pm 1.32$ & $100\%$\\
                           & Cheeks-In     & $8.0 \pm 1.28$ & $94\%$\\
                           & Eyebrow       & $5.7 \pm 0.62$ & $100\%$\\
\bottomrule
\end{tabular}
\end{adjustbox}
\end{table}

\smallskip

\begin{figure}[!t]
\centering
\includegraphics[width=0.99\linewidth]{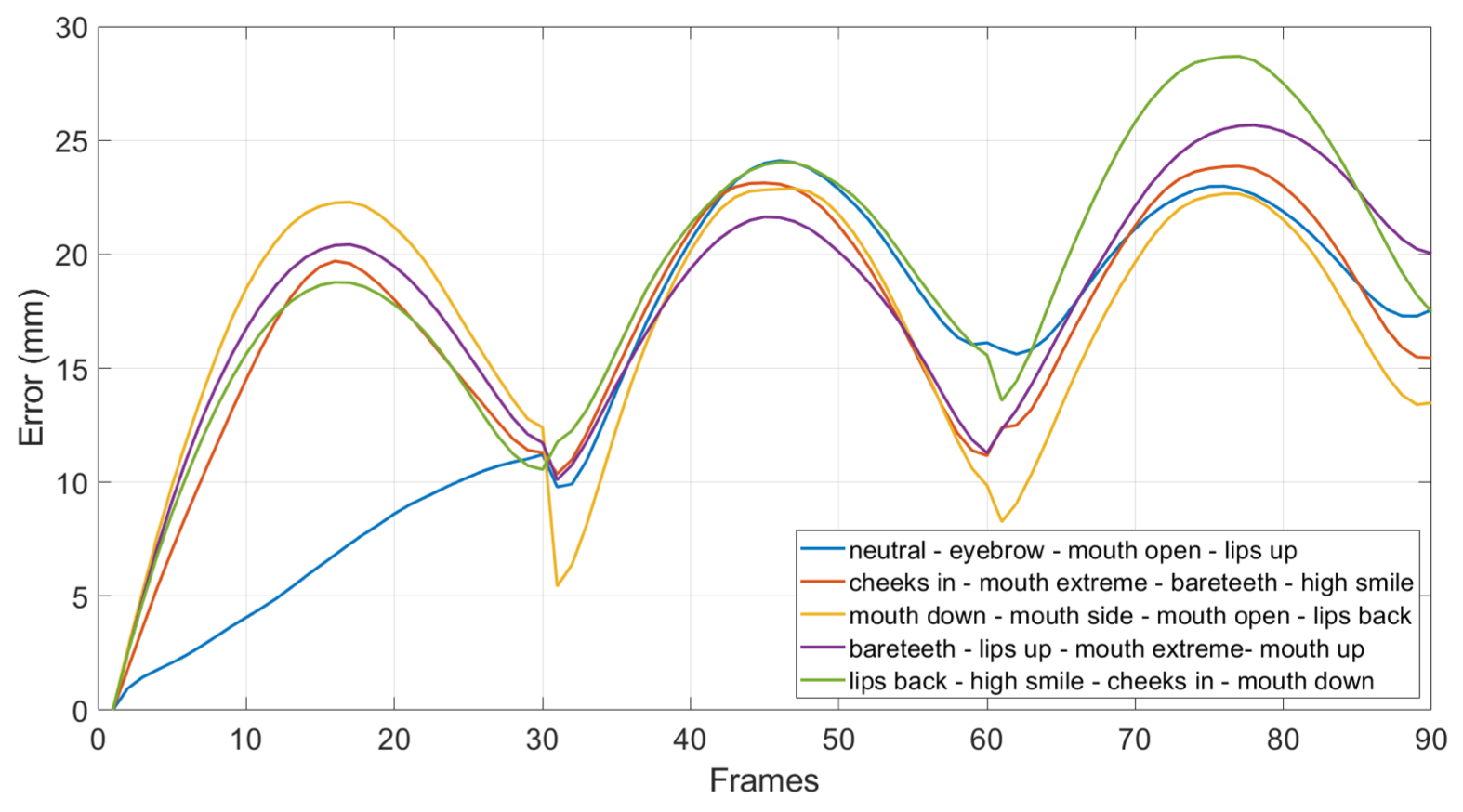}
\caption{Per-frame evolution of the error for 5 sequences composed of 3 transitions each. Each transition is $30$ frames long.}
\label{fig:composed-err}
\end{figure}

\begin{figure*}[!t]
\centering
\includegraphics[width=0.85\linewidth]{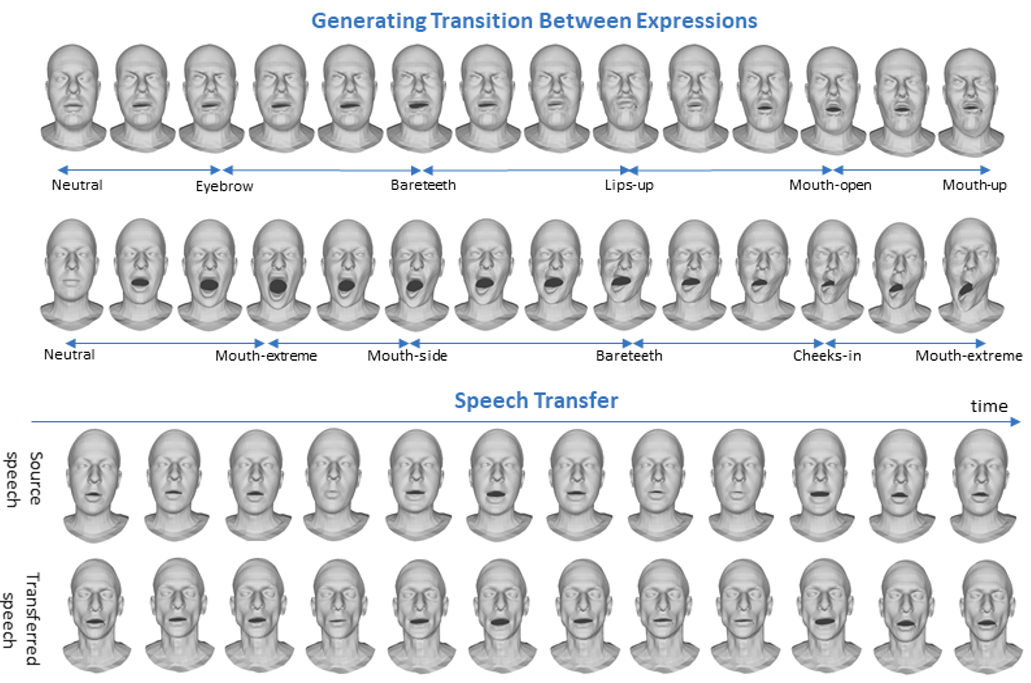}
\caption{\textbf{Applications} -- From top to bottom: \textbf{Transition between expressions}: starting from a neutral face, each row show the transition between five different facial expressions. \textbf{Transfer}: speech transfer from one identity to another. Animated versions can be found in this \href{https://drive.google.com/drive/folders/1_XIu5tx-oFCiHVEW3vs6o_p3H99BwSFe?usp=sharing}{\textcolor{magenta}{link}}. }
\label{fig:applications}
\end{figure*}

\noindent
\textbf{Generating composed sequences}: A novel characteristic of our method is that, even though the length of each transition is fixed to $30$ frames, we are able to generate longer, composed and complex transitions. This is possible as we removed the constraint of starting the animation from a neutral face, and thanks to the SRVF representation, which allowed us to create interpolated transitions from one expression to another. So, it is possible for example to generate a $90$ frames long sequence by composing three transitions, \eg, neutral-bareteeth-eyebrows-lips up. To do so, we generate the sequence incrementally, using as starting landmark configuration for the $(i+1)$-th transition, the ending frame of the $i$-th one. To verify that the model is sufficiently robust to handle the diversity of each generated transition, we generated $64$ samples of $5$ composed sequences of $90$ frames each (3 transitions). The average per-frame error is reported in Figure~\ref{fig:composed-err}. Results show that the error does not significantly propagate across transitions, and remains quite stable even though a slight increasing trend is observed. This is due to the fact by switching from one expression peak to another without getting back to a neutral state, the resulting expression is actually a mix of the two. This is an interesting property, which makes the generated sequences even more natural looking. The lower peaks at frames 30 and 60 are instead due to the fact the training/testing sequences are 30 frames long, so leading to a discontinuity when computing the error. To clarify this aspect, let us consider the sequence ``mouth down-mouth side-mouth open-lips back'' (yellow curve in Figure~\ref{fig:composed-err}): to compute the error from frame 0 to 30, we considered the corresponding transition in the real data; to compute the error for frames 30-60, we instead needed to consider a different transition, though the generated one starts from the last frame of the previous one. Ultimately, the discontinuity is reflected in the errors. Nonetheless, qualitative examples in Figure~\ref{fig:applications} and the supplementary video show the error propagation does not significantly corrupt the output. 

\subsection{Subjective Evaluation}
In this section, we report the results of a user study aimed at assessing the perceived quality of the generated 4D expression sequences. We recruited 23 participants, and asked them to evaluate the quality of the generated data in terms of ``Naturalness'' (NAT) score, ranging from 0 to 10. To obtain consistent results, before starting the questionnaire, users were shown two reference examples for the lower and upper bounds of the score; 0 corresponds to a completely failed animation, while 10 corresponds to a real sequence from the original dataset. In addition, we also asked the users to evaluate the faithfulness of a given animation with respect to the input label (LAB). The latter score also ranges from 0 to 10. Users were shown a total of 28 generated animations, 14 of which included multiple transitions, and 14 only neutral-peak transitions.

The average reported NAT score was of 7.22, while the LAB score of 7.05, indicating users perceived the generated meshes as fairly natural and faithful to the input label. Concerning the neutral-peak sequences, they were generated both using the proposed model, and that of~\cite{Otberdout_2022_CVPR}. The NAT score reported for our model was of 7.39, and 5.82 for the model of~\cite{Otberdout_2022_CVPR}. Our results were perceived as more realistic, possibly thanks to the increased variability and quantity of the training data, whose collection was made possible by augmenting them with interpolated samples.

\subsection{Qualitative Results}
Figure~\ref{fig:applications} shows additional qualitative results. 
In particular, Figure~\ref{fig:applications}~(top) shows two examples of composed sequences generated with our model. These are obtained by generating transitions between different expressions incrementally with Motion3DGAN. S2D-Dec was applied to these motions ultimately obtaining a complex 4D sequence. Figure~\ref{fig:applications}~(bottom), instead, shows some qualitative examples of speech transfer, applied as described in Section~\ref{subsubsec:speech}. The reader can appreciate that our model allows a natural transfer of speech related movements that were completely unseen during training.

\smallskip


\section{Conclusions and Limitations}\label{sect:conclusions}
In this paper, we proposed a novel framework for dynamic 3D facial expression generation. From  a starting 3D face and an expression label indicating the starting and the ending expression, we can synthesize sequences of 3D faces switching between different facial expression. This is achieved by two decoupled networks that separately address the motion dynamics modeling and generation of expressive 3D faces from a starting one. We demonstrated the improvement with respect to previous literature, the high generalization ability of the model to unseen expressions and identities, and showed that using landmarks is effective in modeling the motion of expressions and the generation of 3D meshes. 
We also identified two main limitations: first, our S2D-Dec generates expression-specific deformations, and so cannot model identities. Moreover, while Motion3DGAN can generate diverse expressions including transition between expressions and long composed 4D expressions, the samples are of a fixed length. 

\section{Acknowledgments}
This work was supported by the French State, managed by National Agency for Research (ANR) National Agency for Research (ANR) under the Investments for the future program with reference ANR-16-IDEX-0004 ULNE. This paper was also partially supported by the European Union's Horizon 2020 research and innovation program under grant number 951911 - AI4Media. 
The authors would also like to thank Giulio Calamai for performing part of the experimentation. Most of this work was done when Naima Otberdout was at the University of Lille.

\section{Appendix}

\subsection{Landmarks Configuration}
\begin{figure}[!ht]
\centering
\includegraphics[width=0.7\linewidth]{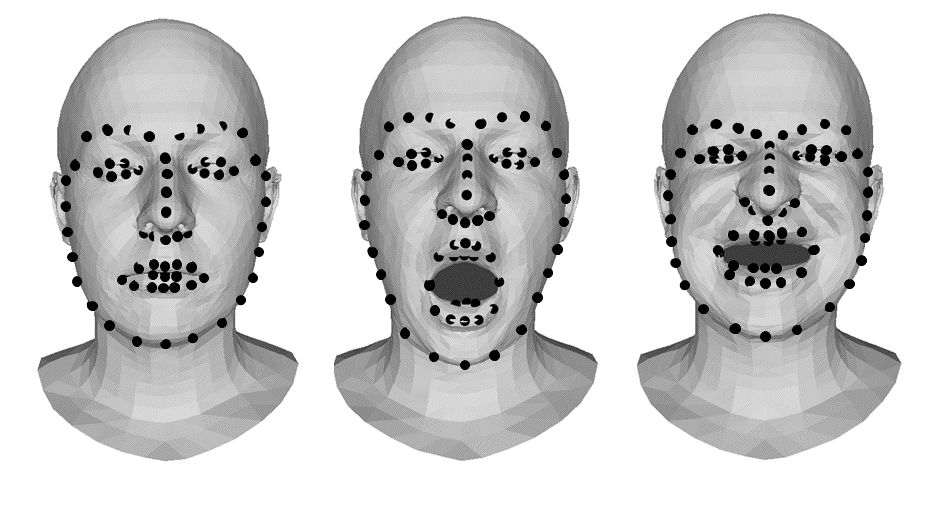}
\centering
\caption{Landmarks configuration used to guide our model.}
\label{fig:landmarks}
\end{figure}

In Figure~\ref{fig:landmarks} we show, for three different expressions, the configuration of landmarks used to guide the generation of the facial expression.

\subsection{Logarithm and Exponential Maps}
In order to map  the SRVF data forth and
back to a tangent space of $\mathcal{C}$, we use the logarithm  $\log _{p}(.)$ and the exponential $\exp _{p}(.)$ maps defined in a given point $p$ by,
\begin{equation}
\begin{split}
\label{Eq:LogSphereLog}
\log _{p}(q) &= \frac{d_{\mathcal{C}}(q,p)}{sin(d_{\mathcal{C}}(q,p))} (q - cos(d_{\mathcal{C}}(q,p))p), \\
\exp _{p}(s)&=cos(\|s\|)p + sin(\|s\|)\frac{s}{\|s\|},
\end{split}
\end{equation}

\noindent
where $d_{\mathcal{C}}(q,p)=\cos^{-1}(\langle q,p \rangle)$ is the distance between $q$ and $p$ in $\mathcal{C}$.

\subsection{Architecture of S2D-Dec}
The architecture adopted for S2D-Dec is based on the architecture proposed in~\cite{Bouritsas:ICCV2019}. S2D-Dec takes as input the displacements of $68$ landmarks illustrated in Figure~\ref{fig:landmarks}. The architecture includes a fully connected layer of size $2688$, five spiral convolution layers of $64$, $32$, $32$, $16$ and $3$ filters. Each spiral convolution layer is followed by an up-sampling by a factor of $4$. 

\begin{figure*}
\centering
\includegraphics[width=0.8\linewidth]{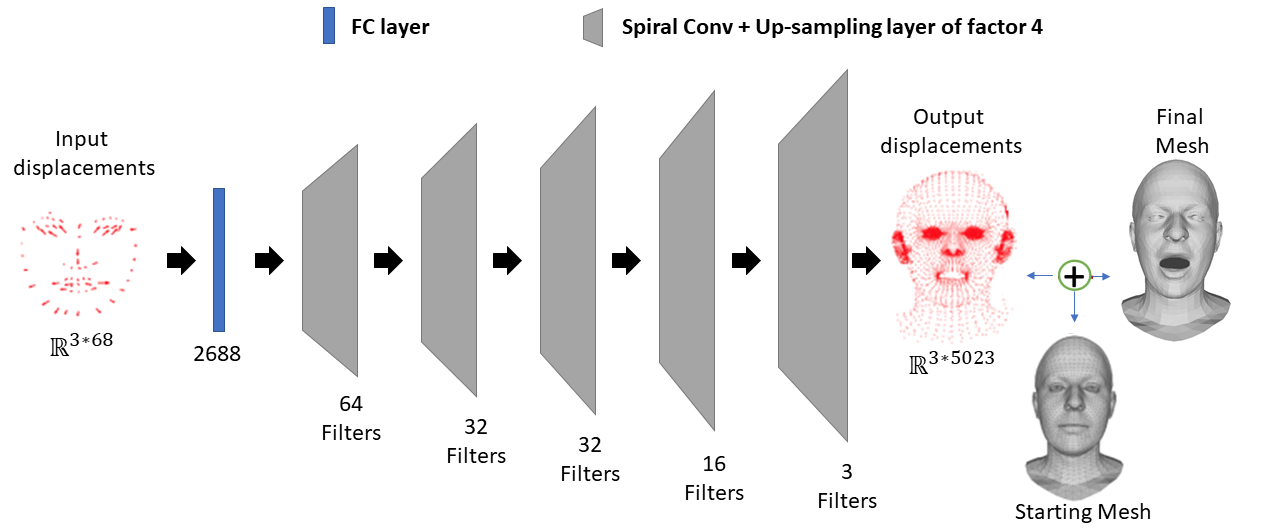}\\
\caption{Architecture of the Sparse2Dense decoder (S2D-Dec). }
\label{fig:DecArchi}
\end{figure*}

\subsection{Ablation Study}

\begin{figure*}
\centering
\includegraphics[width=0.5\linewidth]{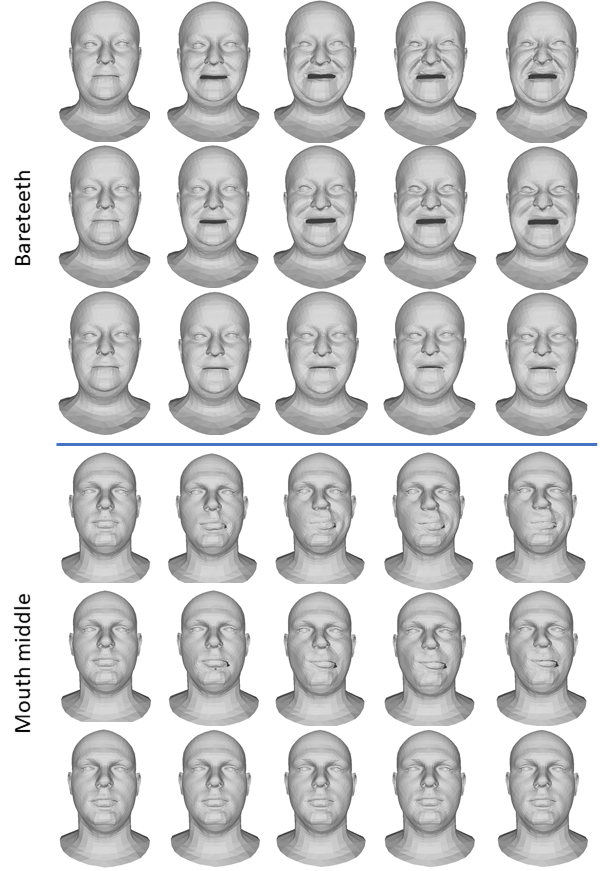}
\centering
\caption{Ablation study: qualitative comparison between ground truth (first row) our
model with (second row) and without (last row) weighted loss.}
\label{fig:ablationsupp}
\end{figure*}

In this section, we report a visual comparison between reconstructions obtained with the standard L1 loss and our proposed weighted L1. Figure~\ref{fig:ablationsupp} clearly shows the effect of our introduced weighting scheme that allows for improved expression modeling. 

\begin{figure*}[!t]
\centering
\includegraphics[width=1\linewidth]{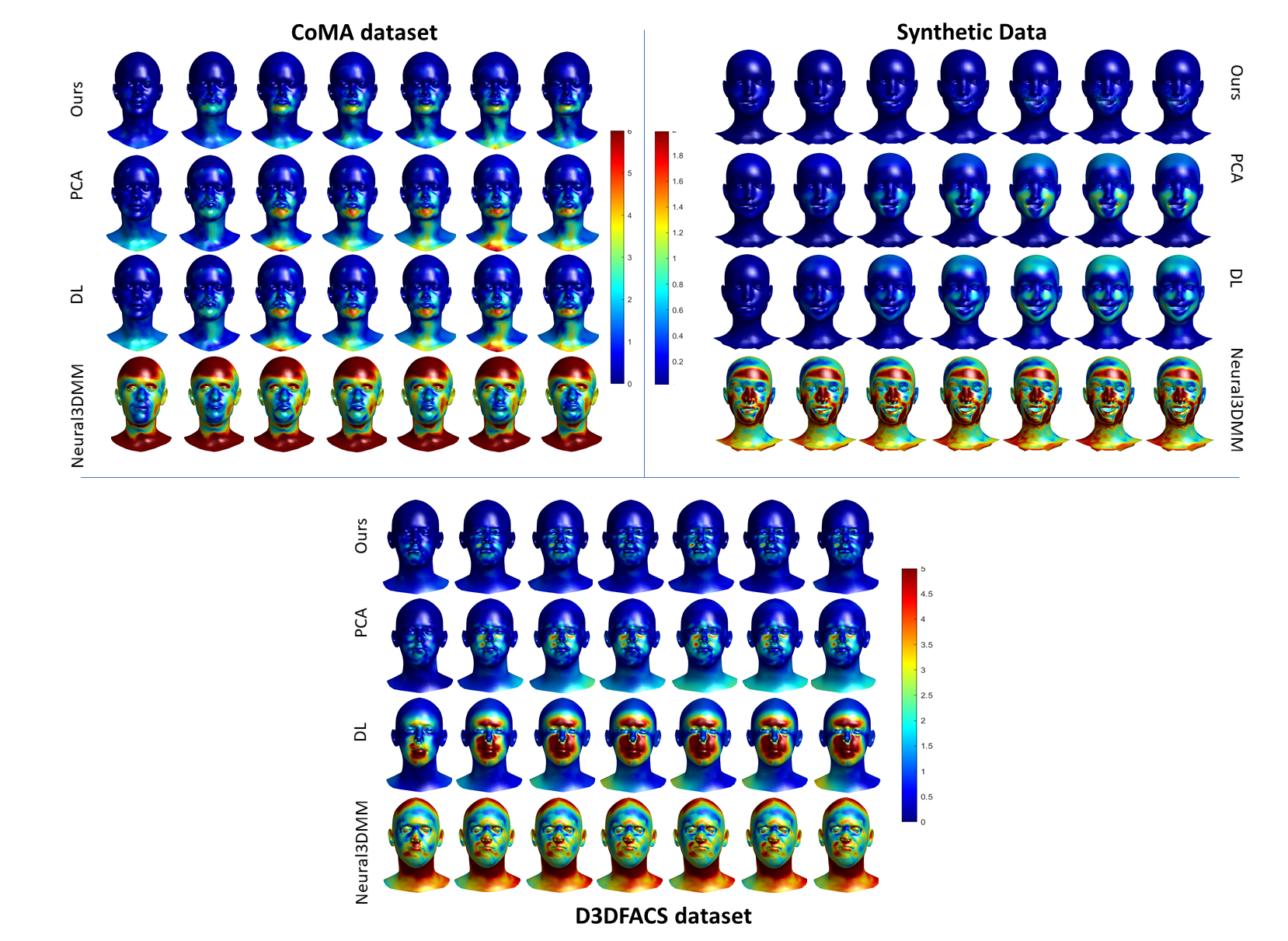}\\
\caption{Temporal evolution of the mesh reconstruction error (red=high, blue=low) from the neutral face to the apex expression of our model and other methods. Examples from three different databases.}
\label{fig:cum-err}
\end{figure*}





%



%

\begin{IEEEbiography}[{\includegraphics[width=1in,height=1.25in,clip,keepaspectratio]{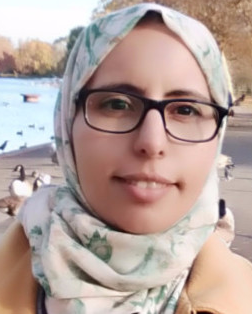}}]
{Naima Otberdout} is currently a Postdoctoral researcher in the University of Lille, France. She received the master’s degree in computer sciences and telecommunication from Mohammed V University, Rabat, Morocco in 2016. She received the Ph.D. degree in computer science from the same university in 2021. Her current research interests include computer vision and pattern recognition with applications to human behavior understanding.
\end{IEEEbiography}
\begin{IEEEbiography}[{\includegraphics[width=1in,height=1.25in,clip,keepaspectratio]{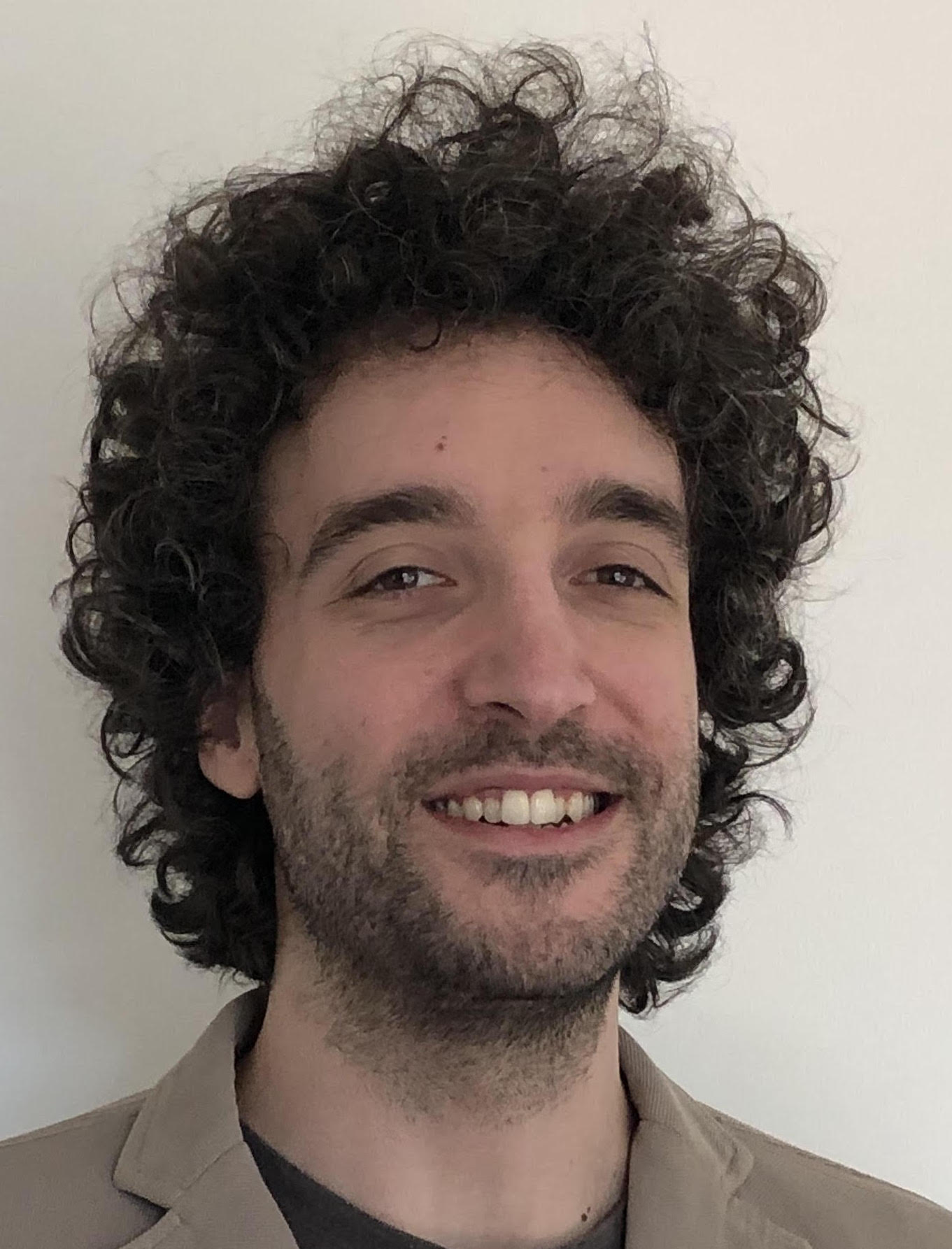}}]
{Claudio Ferrari} is currently assistant professor at the department of Architecture and Engineering of the University of Parma. He received the Ph.D. in Information Engineering from the University of Florence, in 2018. He has been a visiting research scholar at the University of Southern California in 2014. His research interest focus on 3D/4D face and body analysis, human emotion, biometrics and behavior understanding. 
\end{IEEEbiography}

\begin{IEEEbiography}[{\includegraphics[width=1in,height=1.25in,clip,keepaspectratio]{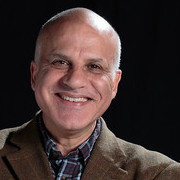}}] {Mohamed Daoudi} is Full Professor of Computer Science at IMT Nord Europe and the lead of Image group at CRIStAL Laboratory (UMR CNRS 9189). His research interests include pattern recognition, shape analysis and computer vision. He has published over 150 papers in some of the most distinguished scientific journals and international conferences. He is/was Associate Editor of Image and Vision Computing Journal, IEEE Trans. on Multimedia, Computer Vision and Image Understanding, IEEE Trans. on Affective Computing and Journal of Imaging. He has served as General Chair of IEEE International Conference on Automatic Face and Gesture Recognition, 2019. Prof. Daoudi is IAPR Fellow.
\end{IEEEbiography}

\begin{IEEEbiography}[{\includegraphics[width=1in,height=1.25in,clip,keepaspectratio]{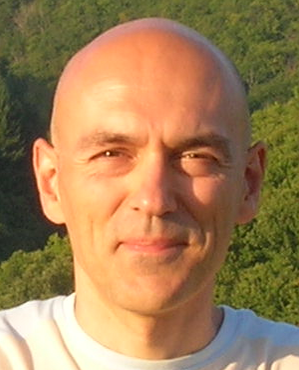}}]{Stefano Berretti} is an Associate Professor at University of Florence, Italy. He has been Visiting Professor at University of Lille, and University of Alberta. His research interests focus on 3D computer vision for face biometrics, human emotion and behavior understanding. On these themes he published more than 200 journals and conference papers. He is an Associate Editor of ACM TOMM, IEEE TCSVT, and of the IET \textit{Computer Vision journal}.
\end{IEEEbiography}

\begin{IEEEbiography}[{\includegraphics[width=1in,height=1.25in,clip,keepaspectratio]{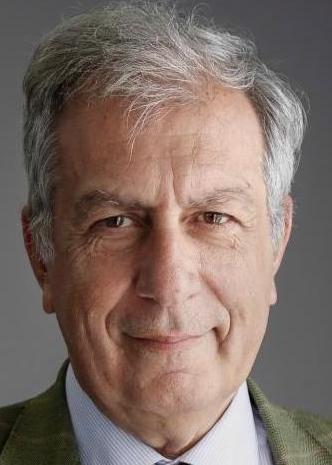}}]{Alberto Del Bimbo} is Full Professor of Computer Engineering at the University of Florence, Italy. His scientific interests include multimedia retrieval, pattern recognition, image and video analysis and human–computer interaction. Prof. Del Bimbo is IAPR Fellow, Associate Editor of several journals in the area of pattern recognition and multimedia, and the Editor-in-Chief of the ACM Transactions on Multimedia Computing, Communications, and Applications. He was also the recipient of the prestigious SIGMM 2016 Award for Outstanding Technical Contributions to Multimedia Computing, Communications and Applications.
\end{IEEEbiography}




\end{document}